# Tuning adaptive gamma correction (TAGC) for enhancing images in low light

Ghufran Alhamzawi
College of Computer Science and
Information Technology
University of Al-Qadisiyah
Diwaniyah, Iraq
cm.post23.10@qu.edu.iq

Alfoudi, Ali Saeed
College of Computer Science and
Information Technology
University of Al-Qadisiyah
Diwaniyah, Iraq
Ali.alfoudi@qu.edu.iq

Suha Mohammed Hadi
Informatics Institute for
Postgraduate Studies
University of Information
Technology and Communications
Baghdad, Iraq
dr.suhahadi@gmail.com

Ali Hakem Alsaeedi
College of Computer Science and
Techniques
Information Technology
Diwaniyah, Iraq
ali.alsaeedi@qu.edu.iq

Amjed Abbas Ahmed
Center for Cyber Security, Faculty of
Information Science and Technology
Universiti Kebangsaan Malaysia
Bangi, Malaysia
Department of Computer Techniques
Engineering, Imam AlKadhum
College (IKC), Baghdad  Iraq
amjedabbas@alkadhum-col.edu.iq

Md. Riad Hassan
Department of Computer Science and
Engineering
Green University of Bangladesh
Baghdad  Iraq
riad@cse.green.edu.bd

Nurhizam Safie Mohd Satar
Center for Cyber Security
Faculty of Information Science and
Technology,
Universiti Kebangsaan
Malaysia (UKM) Bangi, 43600, Malaysia
nurhizam@ukm.edu.my

Waeel Yahya Yasseen
Department of Computer Science
College of Education, Mustansiriyah
University
The Sunni Endowment
Department of Religion Charity Institution
Baghdad,Iraq
waeelyahya@uomustansiriyah.edu.iq

*Abstract*— **Enhancing images in low-light conditions is an important challenge in computer vision. Insufficient illumination negatively affects the quality of images, resulting in low contrast, intensive noise, and blurred details. This paper presents a model for enhancing low-light images called tuning adaptive gamma correction (TAGC). The model is based on analyzing the color luminance of the low-light image and calculating the average color to determine the adaptive gamma coefficient. The gamma value is calculated automatically and adaptively at different illumination levels suitable for the image without human intervention or manual adjustment. Based on qualitative and quantitative evaluation, tuning adaptive gamma correction model has effectively improved low-light images while maintaining details, natural contrast, and correct color distribution. It also provides natural visual quality. It can be considered a more efficient solution for processing low-light images in multiple applications such as night surveillance, improving the quality of medical images, and photography in low-light environments.**

## I.    INTRODUCTION

Low light is a major challenge in computer vision [1]. It reduces the scene's image quality regarding contrast and detail clarity [2]. Imperfect lighting increases noise and loss of detail in dark areas [3]. Low light image enhancement plays an important role in human life because it enters into many fields, the most important of which are the military [4], security, health, night photography[5], and smart driving fields [6]. Low light image enhancement techniques aim to restore lost information and enhance lighting without overdoing it, in addition to achieving a balance between removing noise, enhancing contrast, restoring details, and improving colors so that images are closer to natural images [7].

Many image processing techniques have been used to improve the quality of low-light images, such as histogram equalization methods [8], Retinex theory [9], Dehazing [10], gamma correction (GC) [11] and techniques based on deep learning [12][13], that require huge training data and high computational resource consumption.

An approach proposed in this paper uses tuning adaptive gamma correction (TAGC), a technique based on dynamically adjusting the illumination distribution within an image based on local analysis of brightness levels [14]. Details in low-light areas are improved without distortion or loss of information in bright areas. It allows visual improvements to images without brightening bright parts or amplifying noise in dark areas [15]. Fig.1. shows the intensity distribution of low light and enhanced image by tuning adaptive gamma correction (TAGC).

The contributions that the proposed model aims to achieve are

1.   Increasing the processing speed reduces the computational and space complexities.

2. Enhancing the illumination in the image is very similar to normal illumination and restoring the image with high color accuracy.

3. Processing images with mixed illumination and choosing the appropriate basic parameters, such as the gamma value

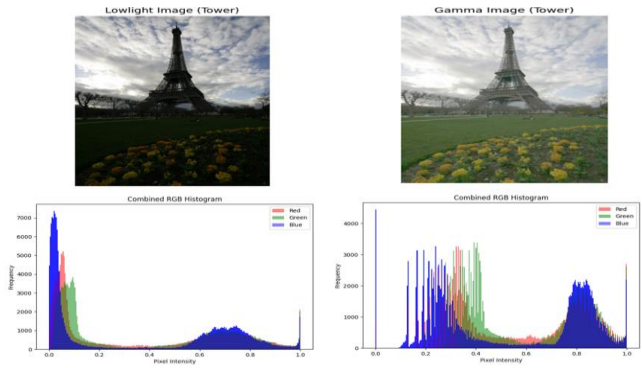

Fig. 1. the intensity distribution of low light and enhanced image by tuning adaptive gamma correction.

## II. RELATED WORTS

There are many techniques for enhancing low light images.

Al-Wadud et al. [16] proposed a dynamic histogram equation (DHE) method for contrast enhancement of images. In this method, the input histogram is divided into some sub-histograms. After passing through the dynamic histogram equation, each sub-histogram can occupy a particular gray level range in the enhanced output image [17]. This method may be complex to implement because it requires subdividing the histogram and performing segmentation tests.

LI et al. [18]presented a fusion method to produce high contrast for bright areas and improve the visibility of dark parts for low-light images. The authors relied on defining a pair of complementary gamma functions (PCGF) [19]. They proposed a new sharpening technique that further enhances the contrast and detail of low-light images. They implemented a simple fusion technique to merge the value components of the enhanced image into the HSV space generated by sharpening and PCGF. The method relies on an initial image enhancement using complementary gamma functions, which may affect performance when this step is ineffective [20]. Complementary gamma functions may require fine-tuning of parameters, making performance dependent on user experience.

Xiu Ji et al. [21]presented a method based on Retinex theory to enhance illumination and improve image details. An adaptive color-balancing technique dealt with color differences in low-light images. The captured image was converted from RGB to HSV space, and the illumination and reflectance components were accurately extracted using a multi-scale Gaussian function combined with Retinex theory [22]. The light components were separated into regions with high and low illumination levels and enhanced. Each block region of the image was weighted and combined. They applied the detail enhancement algorithm to improve the image details further. The performance of this method needs to improve in low-light areas [23]. The low-light images suffered from over-saturation, indicating that the method may fail to achieve the desired balance.

Wang et al. [24] presented an image enhancement technique based on adaptive local gamma transform and color compensation inspired by the low-light image enhancement (LLIE) illumination inversion model [25]. This technique transforms the original image into the YUV color space [26]. The Y component is estimated using a fast vector filter. The local gamma transform function improves the image's brightness by adaptively adjusting the parameters. The color compensation mechanism and linear stretching strategy improve the image's dynamic range [27]. In this technique, the YUV transform and light adjustment process may lead to the loss of fine details in the images, especially in areas with very low light [28].

## III. THE PROPOSED MODEL TUNING ADAPTIVE GAMMA CORRECTION (TAGC)

The proposed model (TAGC) aims to produce a clear, enhanced image free of noise and unwanted colors (artifacts) with clear details and a balanced color and lighting distribution close to the natural image [29]. The model analyzes the image's brightness and calculates the average color to determine the adaptive gamma coefficient. The gamma value is calculated in an automatic adaptive manner with different lighting levels and is suitable for the image without human intervention or the need for manual adjustment [30].

The proposed model includes four steps.

Step1. Calculating the luminance factor of the image

The brightness level of each pixel in the image is calculated using a mathematical equation.

Step2. Calculating the overall average color of the image

The average values of the color channels (red, green, blue) are calculated using a mathematical equation.

Step3. Calculating the adaptive gamma coefficient

The ideal gamma value is calculated using a mathematical equation based on the color luminance value and the overall average color value of the image.

Step4. Applying adaptive gamma correction

Adaptive gamma correction improves the low-light image by applying a mathematical equation to the value of each pixel.

### A. Luminance factor (L)

Luminance is a measure of brightness in an image. It expresses the intensity of light reflected or emitted from each pixel in the image [31]. It is calculated based on the combination of the three-color channels RGB (red, green, blue) and based on the sensitivity of the human eye; a different weight is given to each channel. The Luminance factor(L) is calculated using equation 1 [32].

$$L = 0.2126 I_R^c + 0.7152 I_G^c + 0.0722 I_B^c \qquad (1)$$

Where $I_R^c$, $I_G^c$, and $I_B^c$ refer to the mean's channels of red, green, and blue in the given order.

### B. Average color factor($\breve{\mu}$)

The average color factor is the sum of the average values of each three-color channel RGB (red, green, blue). It is calculated by Equation 2 [33].

$$\breve{\mu} = \sum_{i=1}^{n} I_i^c / 3 \qquad (2)$$

Where $I_i^c$ refers to an average of three channels R, G and B, c is 1,2,3.

### C. Adaptive gamma coefficient

The adaptive gamma coefficient is calculated for each pixel based on the luminance factor(L)

and average color factor $(\acute{\mu})$ using Equation 3 [33].

$$\gamma = \gamma_c + [(0.5 - L) \times (1 - \acute{\mu})] - 2L \qquad (3)$$

Where $\gamma_c$ is the control parameter in this model equal to 5.

### D. Appling adaptive gamma correction

Equation 4 applies adaptive gamma correction to each pixel in the image to obtain the final enhanced image using tuning adaptive gamma correction (TAGC).

$$I_{E'} = A \times (I_x)^{2/\gamma} \qquad (4)$$

Where $I_x$ refers to the input image, $I_{E'}$ refers to the enhanced image by adaptive gamma corrected (TAGC), and A is a constant number in this model equal to 1.

The overall algorithm of the proposed tuning adaptive gamma correction (TAGC) model is introduced in Algorithm A.

### Algorithm A: TAGC Model

1. Input: Low light image ($I_x$)

2. Output: Enhanced image ($I_E$)

3. Calculating Luminance to the input image ($I_x$) (Equation 1)

4. Calculating average color to the input image ($I_x$) (Equation 2)

5. Calculating adaptive gamma coefficient (Equation 3)

6. Apply adaptive gamma correction to the input image ($I_x$) (Equation 4)

7. Return enhanced image ($I_E$)

## IV. RESULT DISCUSSION

To evaluate the proposed model's performance, it was tested on two datasets, paired and unpaired. The results were compared with three state-of-the-art algorithms: IAGC [34], LGMS [35], and LIEW [36].

### A. Datasets description

The proposed model was tested on four datasets. Two of the datasets are paired: the LOL dataset [37] contains 485 pairs of images, and the Brightening Train dataset contains 1000 pairs of images. The other two datasets are unpaired: the DICM dataset [38] contains 69 images, and the LIME dataset [39] contains 10 images.

### B. Quantitative evaluation

To perform the quantitative comparison of the enhanced image from the paired datasets lol and Brightening Train, three metrics based on the availability of the ground truth were used: PSNR [26] (Peak to Signal to Noise), SSIM [40] (structural similarity index) and FSIM [41] (Feature Similarity Index).

The Natural Image Quality Evaluator (NIQE) also quantitatively compared the enhanced images from the unpaired datasets DICM and LIME [42].

The results in Table I. show that the proposed TAGC model is the most efficient and balanced among all the models. It excels in preserving visual features and structural details of the image and provides excellent illumination enhancement compared to competing algorithms. The proposed (TAGC) model achieved the highest SSIM value on two reference datasets, LOL and Brightening Train, with values of 0.671 and 0.764, respectively. This means that the proposed model (TAGC) can recover the structural details of the original image better than other models [43]. The proposed model achieved an outperforming of the compared algorithms by FSIM↑ criterion with a value of 0.894 on the Brightening Train dataset and a value of 0.873 on the LOL dataset. That indicates the ability of the (TAGC) model to recover the fine visual details and structural edges of the image more accurately than other models [44]. The proposed model (TAGC) also achieved an outperforming of the compared algorithms by PSNR↑ criterion with a value of 17.796 on the Brightening Train dataset and a value of 14.551 on the LOL dataset. That indicates the ability of the (TAGC) model to enhance images while maintaining their original quality with the least amount of distortion or information loss [45]. The performance is evaluated in Table II. using the NIQE metric (lower values are assumed to indicate better performance, such as error measurement or quality loss). The proposed model (TAGC) achieved the lowest value in all cases, as it obtained 3.713 on the DICM dataset and 3.877 on the LIME dataset [46]. That indicates superiority in producing improved images with less distortion and noise than previous models, which leads to preserving image details and quality more efficiently.

TABLE I. COMPARE THE PROPOSED MODEL'S (TAGC) RESULTS WITH THE SET OF ALGORITHMS ON TWO BENCHMARK DATASETS, LOL AND BRIGHTENING TRAIN, USING PSNR, SSIM, LPIPS, AND FSIM METRICS.

| Algorithm | Dataset | | | | | |
| | LOL | | | Brightening Train | | |
| | PSNR↑ | SSIM↑ | FSIM↑ | PSNR↑ | SSIM↑ | FSIM↑ |
|---|---|---|---|---|---|---|
| IAGC [34] | 11.260 | 0.468 | 0.864 | 14.056 | 0.663 | 0.861 |
| LGMS [47] | 15.834 | 0.475 | 0.847 | 16.943 | 0.747 | 0.902 |
| LIEW [36] | 12.678 | 0.638 | 0.880 | 16.943 | 0.747 | 0.902 |
| Proposed model (TAGC) | 14.551 | 0.671 | 0.873 | 17.796 | 0.857 | 0.915 |

| Algorithm | Dataset | | Average |
|---|---|---|---|
| | *DICM* | *LIME* | |
| IAGC [34] | 4.007 | 3.948 | 3.977 |
| LGMS [47] | 3.941 | 4.384 | 4.162 |
| LIEW [36] | 3.804 | 4.096 | 3.95 |
| Proposed model (TAGC) | 3.713 | 3.877 | 3.795 |

## C. Qualitative evaluation

The images were evaluated visually, as shown in Fig. 2. and Fig.3. The images were examined in different models, and it was observed that the proposed model (TAGC) produces enhanced images with balanced brightness and high contrast while preserving fine details [48]. The model IAGC shows a slight improvement in lighting and suffers from color distortions, unwanted color (artifacts), and loss of fine details in dark areas. The model LGMS enhances lighting, but the enhanced images had dark and unnatural colors compared to the original image, which negatively affected the realism of the images after processing [49]. The model LIEW achieved acceptable improvement, but it could not provide a complete restoration of the original colors and details, as the enhanced images suffered from a clear decrease in contrast, which led to the loss of some details in the images [50].

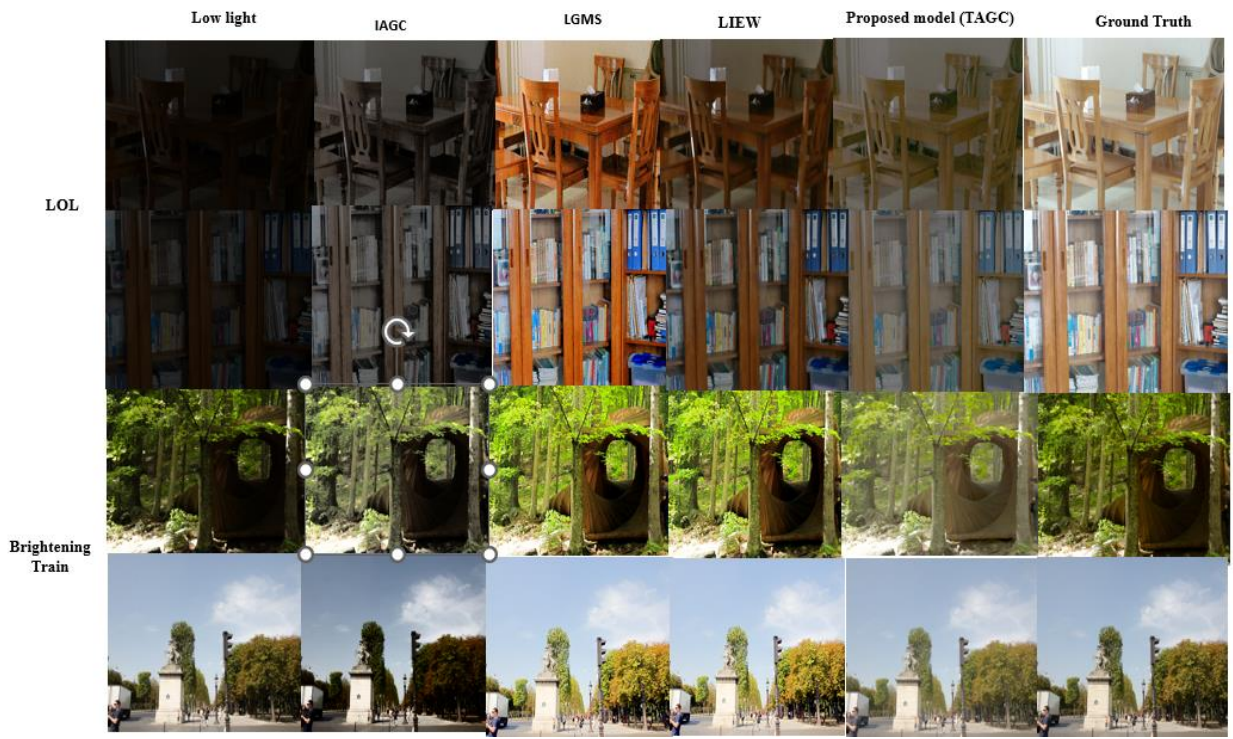

Fig.2. Comparison of some images generated by the proposed model with images generated by a set of algorithms on the two paired datasets LOL dataset and Brightening Train dataset .

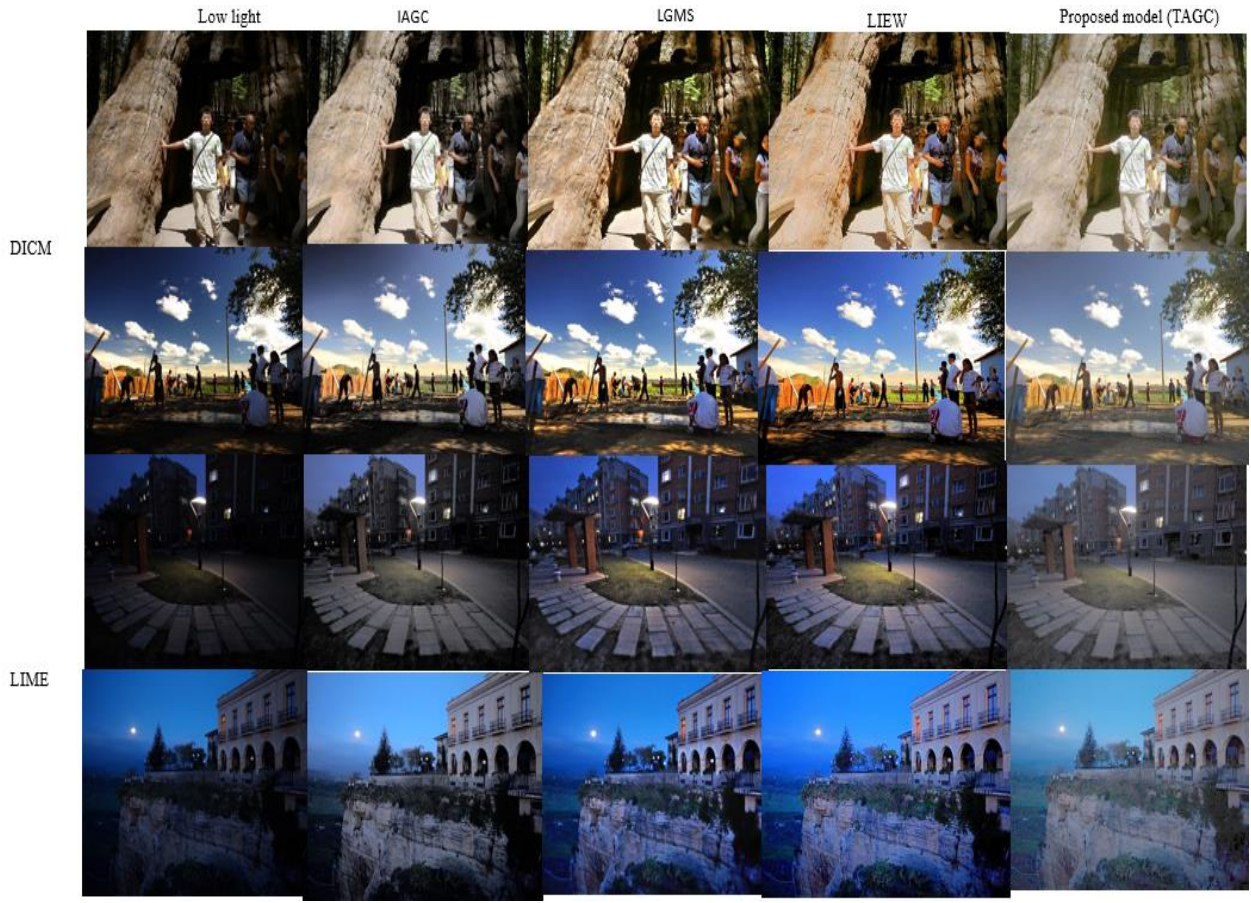

Fig.3. Comparison of some images generated by the proposed model with images generated by a set of algorithms on the two unpaired datasets DICM dataset and LIME dataset.

## V. CONCLUSION

Image enhancement techniques in low-light conditions aim to enhance the illumination in a balanced manner without affecting the fine details in the image. Most image enhancement techniques suffer from problems of noise, unwanted color artifacts, and weak or excessive illumination, in addition to computational complexity and time. The proposed model (TAGC) was adaptive to the image's brightness, as the color density distribution of the enhanced images was closer to natural colors. It is suitable for enhancing mixed-light images. It ensures an enhanced image with clear details and is free of noise. The proposed model (TAGC) does not need training data. It is suitable for devices with limited capabilities. The proposed model (TAGC) achieved good results, as the NIQE rate was 3.795 on unpaired data sets and achieved a reclassification ratio of 0.857 to SSIM. The proposed model (TAGC) can be combined with other enhancement techniques to improve enhanced images' accuracy and obtain better future results.